# Facial Expression Recognition Using a Hybrid CNN–SIFT Aggregator

Mundher Al-Shabi, Wooi Ping Cheah, Tee Connie

Faculty of Information Science and Technology, Multimedia University, Melaka, Malaysia

**Abstract.** Deriving an effective facial expression recognition component is important for a successful human-computer interaction system. Nonetheless, recognizing facial expression remains a challenging task. This paper describes a novel approach towards facial expression recognition task. The proposed method is motivated by the success of Convolutional Neural Networks (CNN) on the face recognition problem. Unlike other works, we focus on achieving good accuracy while requiring only a small sample data for training. Scale Invariant Feature Transform (SIFT) features are used to increase the performance on small data as SIFT does not require extensive training data to generate useful features. In this paper, both Dense SIFT and regular SIFT are studied and compared when merged with CNN features. Moreover, an aggregator of the models is developed. The proposed approach is tested on the FER-2013 and CK+ datasets. Results demonstrate the superiority of CNN with Dense SIFT over conventional CNN and CNN with SIFT. The accuracy even increased when all the models are aggregated which generates state-of-art results on FER-2013 and CK+ datasets, where it achieved 73.4% on FER-2013 and 99.1% on CK+.

**Keywords:** Facial Expression Recognition, Dense SIFT, CNN, SIFT.

## 1    Introduction

Automatic facial expression recognition is an interesting and challenging problem which has important applications in many areas like human-computer interaction. It helps to build more intelligent robots which has the ability to understand human emotions. Many other real-world applications such as call center and interactive game development also benefit from such intelligence.

Ekman in early 1970s showed that there are six universal emotional expressions across all cultures, namely disgust, anger, happiness, sadness, surprise and fear [3]. These expressions could be identified by observing the face signals. For example, a smile gesture by raising the mouth corners and tightening the eyelids is a signal of happiness.

Due to the importance of facial expression recognition in designing human–computer interaction systems, various feature extraction and machine learning algorithms have been developed. Most of these methods deployed hand-crafted features followed by a classifier such as local binary pattern with SVM classification [20], Haar [24],

SIFT[1], and Gabor filters with fisher linear discriminant [12], and also Local phase quantization (LPQ) [23].

The recent success of convolutional neural networks (CNNs) in tasks like image classification [5] has been extended to the problem of facial expression recognition [9]. Unlike traditional machine learning and computer vision approaches where features are defined by hand, CNN learns to extract the features directly from the training database using iterative algorithms like gradient descent. CNN is usually combined with feed-forward neural network classifier which makes the model end-to-end trainable on the dataset.

Like ordinary neural network, CNN learns its weights using back-propagation algorithm. It has two main components namely local receptive fields and shared weights. In local receptive fields, each neuron is connected to a local group of the input space. The size of this group of the input space is equal to the filter size where the input space can be either pixels from the input image or features from the previews layer. In CNN the same weights and bias are used over all local receptive fields. These two components, although make CNN runs faster, but are prone to over-fitting as the same weights are applied to the entire image.

In most cases, CNN requires a lot of training data to generalize well. The availability of large datasets and cheap computational power offered by GPU increase the popularity of CNN. However, this is not the case in facial expression recognition where the datasets are limited. While Scale Invariant Feature Transform (SIFT) [15] and other hand-crafted methods provide less accurate results than CNN[2, 11, 13], they do not require extensive datasets to generalize. Nonetheless, the modeling capacity of the hand-crafted methods are limited by the fixed transformations (filters) that remain the same for different sources of data. In this paper, we propose a hybrid approach by combining SIFT and CNN to get the best of both worlds. Both regular SIFT and Dense SIFT are investigated and combined with CNN model. Fig. 1 shows an overview of the proposed approach. The raw image passes through the CNN layers before combined with either SIFT or Dense SIFT features. Both SIFT or Dense SIFT models are trained and evaluated separately. The SIFT and Dense SIFT features are merged with the CNN features at the final stage. Unlike other works, the CNN and SIFT/ Dense SIFT features are combined during the training and testing phases. Moreover, the CNN features and fully-connected layers for the SIFT features are trained at same time which makes this work unique. The existence of the SIFT features during the CNN training help the CNN to learn different features representation from SIFT and make CNN and SIFT compliment each other. The contributions of this paper are two-fold: 1) we investigate the impact of combining SIFT and Dense SIFT with CNN feature to increase the performance of facial expression recognition, and 2) designing a novel classifier for facial expression recognition by aggregating various CNN and SIFT models that achieves a state of art results on both FER-2013 and CK+ datasets.

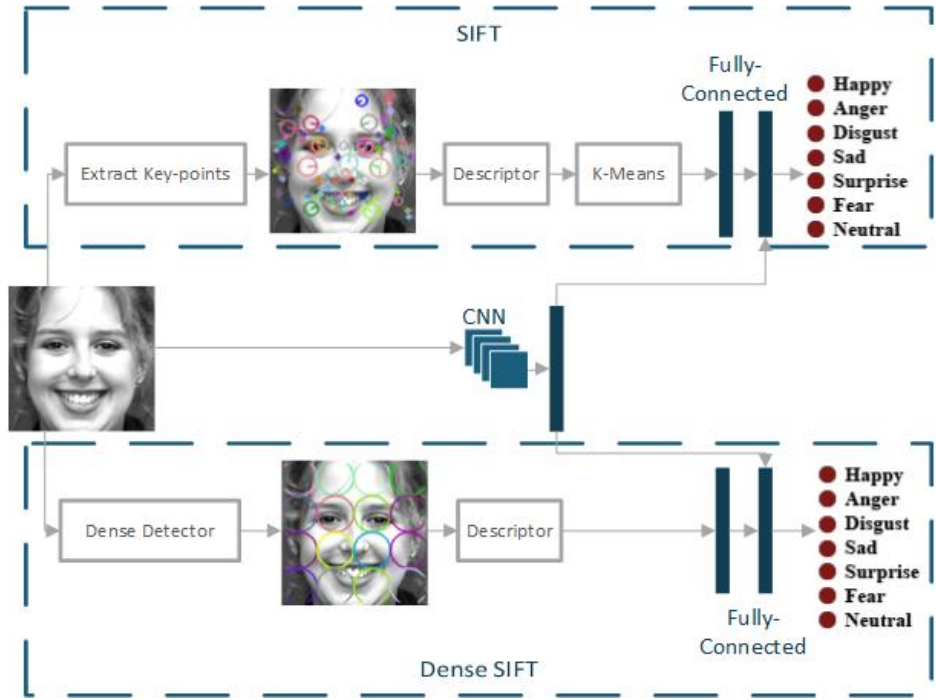

**Fig. 1.** Overview of the proposed methods

## 2 Related Work

Automatic facial expressions recognition (FER) has been an active research in the computer vision field. Facial expression and emotion recognition with hand-crafted feature extractors were reported in [2]–[5]. Many works have also applied convolution neural network in facial expression recognition. In [8] the authors analyzed the features learnt by neural network and showed that neural network could learn patterns from the face images that correspond to Facial Action Units (FAUs). He proposed to ignore the biases of the convolutional layers which gave him an accuracy of 98.3% on the CK+ dataset. The winner of FER-2013 challenge[22] used a CNN layer followed by a linear one-vs-all SVM. Instead of minimization of cross-entropy loss like vanilla CNN, he minimized a margin-based loss with standard hinge loss. The method achieved 71.2% accuracy on a private test. Another study applied deeper neural network by constructing four inception layers after two ordinary convolution layers [18]. Models based on transfer features from pre-trained deep CNN have also been proposed [19, 25]. The importance of pre-processing on increasing the accuracy of the FER model was heavily studied by [14]. Different methods were used to increase the number of examples through rotation correction and intensity normalization.

More recently, ensemble methods such as Bagging or Boosting are applied in facial expressions recognition. Several popular approaches such as [7] used CNN to analyze the videos and deployed deep belief net to capture audio information. The top performing models were then fused as a single predictor. Besides, multiple CNN models were combined via learnable weights by minimizing the hinge loss [26]. The winner of EmotiW2015 [8] trained multiple CNNs as committee members and combined their decisions via construction of a hierarchical architecture of the committee with exponentially-weighted decision fusion. The network architecture, input normalization, and random weight initialization were changed to obtain varied decisions for deep CNNs.

A work reported in [27] extracted fixed number of SIFT features from facial landmarks. A feature matrix consisting of the extracted SIFT feature vectors was used as input to CNN. Another mixture of SIFT and deep convolution networks was proposed in [21]. The authors used dense SIFT, LBP and CNN extracted from AlexNet. The features were trained by linear SVM and Partial least squares regression and the output from all classifiers were combined using a fusion network. Our proposed method is different from [21] in which we use a custom architecture where CNN and the fully-connected layers after the SIFT extractor are trained together. We preferred to design CNN network from scratch other than re-using other architecture to ensure the suitability of CNN features to detecting the expression on the face. Furthermore, the proposed architecture has less complexity and is much smaller in terms of free-parameters which make the model lightweight. It is favorable to make the models lightweight especially when more than model is needed for the aggregator. Additionally, merging the SIFT feature inside the CNN model reduce the possible redundant in features between SIFT and CNN as the later will try to learn different representation from SIFT.

## 3    Pre-Processing

We standardize the size of all images to 48x48 pixels. To make the model more robust to noise and slight transformations, data augmentation is employed. Each image is amplified ten times using different linear transformations. The transformations include horizontal flip, rotation with a random angle between (-30, 30), skewing the center area, and zooming at four corners of the image. Some results of applying the transformations are depicted in Figure 2. Finally, all images are normalized to have zero mean and unit variance.

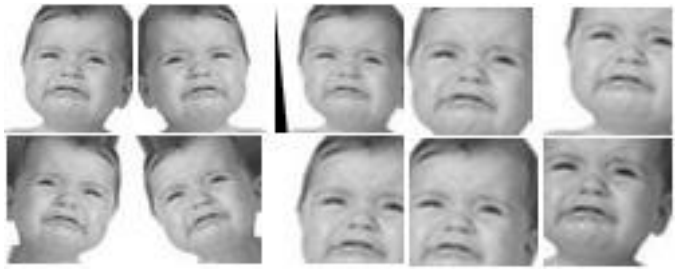

**Fig. 2.** The ten type of image transformations

## 4 Deep CNN Architecture

We built a custom CNN network architecture that is designed from scratch. The network consists of six convolution layers, three Max-Pooling layers, followed by two dense fully connected layers. Each time Max-Pooling is added, the number of the next convolution filters doubles. The number of convolution filters are 64, 128, and 256, respectively. The window size of the filters is 3x3. Max pooling layers with a stride of size 2x2 is placed after every two convolutional layers. Max-Pooling is used to summarize the filter area which is considered as a type of non-linear down-sampling. Max-Pooling is helpful in providing a form of translation invariance and it reduces the computation for the deeper layers.

To retain the spatial size of the output volumes, zero-padding is added around the borders. The output of the convolution layers is flatted and fed to the dense layer. The dense layer consists of 2048 neurons linked as a fully connected layer.

Each of the Max-Pooling and dense layers is followed by a dropout layer to reduce the risk of network over-fitting by preventing co-adaptation of the feature extractor. Finally, a softmax layer with seven outputs is placed at the last stage of the network. To introduce non-linearity for CNN, we used Leaky Rectifier Linear Unit (Leaky ReLU)[17] as follows:

$$f(x) = \max(x, \frac{x}{20}) \qquad (1)$$

where the threshold value 20 is selected using the FER-2013 validation set. Leaky ReLU is chosen over ordinary ReLU to solve the dying ReLU problem. Instead of giving zero when x < 0, leaky ReLU will provide a small negative slope. Besides, its derivatives is not zero which make the network learns faster than ReLU. A categorical cross-entropy method is used as the cost function and is optimized using Adam [10] which is an adaptive gradient-based optimization method. All the hyper-parameters of the network such as the size of each layer are validated and fine-tuned against the FER-2013 validation set.

## 5 SIFT AND BAG OF KEY-POINTS

SIFT [9] is used to extract the key-points from the facial images. After locating the key-points, direction and magnitude of the image gradient are calculated using key-point neighboring pixels. To identify the dominant directions, the gradient histogram is established as shown in Fig. 3. Finally, the SIFT descriptor is determined by partitioning the image into 4x4 squares. For each of the 16 squares, we have a vector length of 8. By merging all the vectors, we obtain a vector of size 128 for every key-point.

In order to use the key-point descriptors in classification, a vector of fixed-size is needed. For this purpose, K-means is used to group the descriptors into a set of clusters. After that, a bag of key-points is formed by calculating the number of descriptors that have been included in each cluster. The resulting feature vector has a size of K. After several trials, an optimal size of K = 2048 is found.

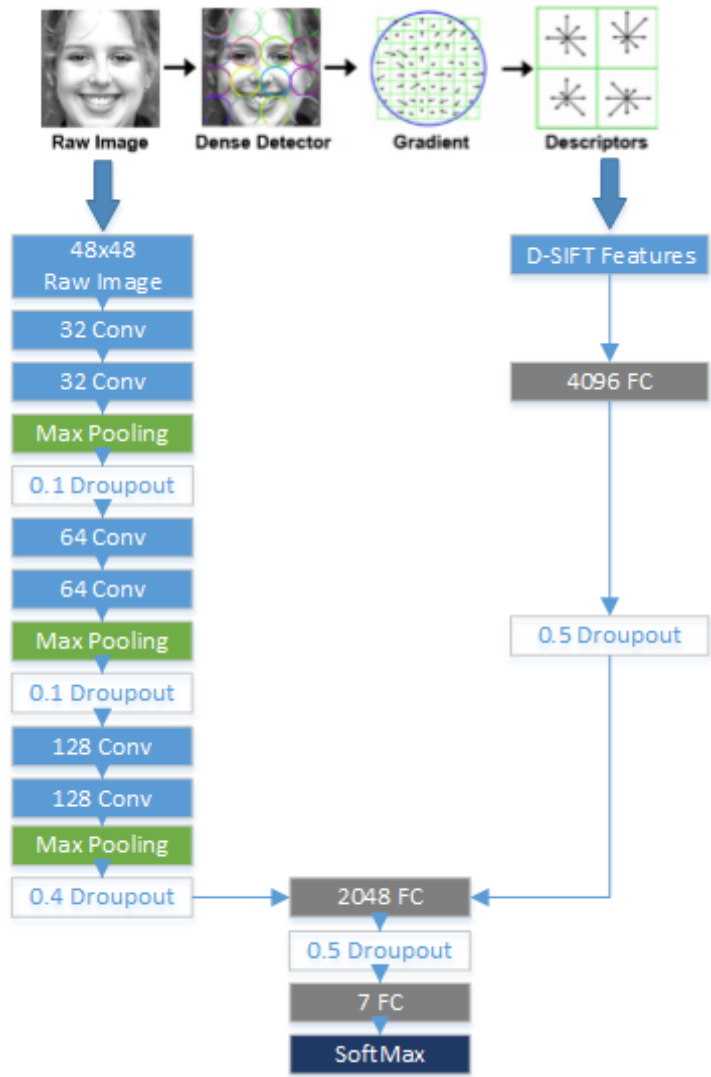

**Fig. 3.** CNN with Dense SIFT

The K-vectors found are passed to a fully-connected layer of size 4096 followed by a dropout. The weights of the fully-connected layer is regularized by l2 norm with value 0.01. Finally the output is merged with the CNN model.

In contrast, dense SIFT does not require extraction of key-points from the facial image. Dense SIFT divides the image into equal region of pixels. Each region has a size of 12x12 pixels which yields 16 regions for an image. The SIFT descriptor is ran over all the 16 regions and each region is described by 128 features. This yields a total of 2048 features for the whole image.

To increase the accuracy of the model, the outputs from CNN only, CNN with SIFT, and CNN with Dense SIFT are aggregated using average sum. The probability of an input image $x$ containing an expression $e$ is:

$$P(e|x) = \frac{A(e|x) + B(e|x) + C(e|x)}{3} \tag{2}$$

where $A$, $B$ and $C$ denote the CNN only model, CNN with SIFT model, and CNN with dense SIFT model, respectively. As each model has a SoftMax layer as the last layer, the output is confined in the range from 0 to 1. The best match expression is the one which yields the highest probability,

$$Y(x) = \underset{e \in Expressions}{\arg\max} \ P(e|x) \tag{3}$$

## 6    Experimental Results

We tested our models on the FER 2013 and Extended Cohn-Kanade (CK+) datasets. The first is FER-2013 which has very wild facial expression images. The second is the standard CK+ which has a tiny number of samples. The FER-2013 dataset was presented in the ICML 2013 Challenges in Representation Learning [4]. The dataset was retrieved using Google image search API. OpenCV face recognition components were used to obtain bounding boxes around each face. The incorrectly labeled images were rejected by human.

On the other hand, the CK+[16] is a lab controlled dataset which consists of 327 images from 123 subjects. Each of the image is assigned one of seven expressions: anger, contempt, disgust, fear, happy, sad, and surprise. To make our experiments compatible with other works like [14], [15], [23] and also the FER-2013 dataset, the contempt examples are removed. So we trained our models on 309 images from the other six expressions.

The FER 2013 dataset contains 28709 training images: 3589 validation (public) and 3589 test (private) divided into seven types of expression Angry, Disgust, Fear, Happy, Sad, Surprise and Neutral. Due to label noise, the human accuracy of this data is 68%. Table 1 shows the distributions of the expressions along the FER-2013 and CK+ datasets. The FER-2013 has more samples than CK+ in all categories. Happiness is the most frequent expression in the dataset. The rest of emotion labels except Disgust has quite similar distribution. Figure 4 shows examples of CK+ and FER-2013 datasets. The images in the FER-2013 dataset is by far more challenging as every image has a different pose.

**Table 1.** Distribution of the emotion labels in the datasets

| Expressions | FER-2013 | CK+ |
|-------------|----------|-----|
| Anger       | 4953     | 45  |

| Disgust | 547 | 59 |
|---------|------|----|
| Fear | 5121 | 25 |
| Happiness | 8989 | 69 |
| Sadness | 6077 | 28 |
| Surprised | 4002 | 83 |
| Neutral | 6198 | 0 |

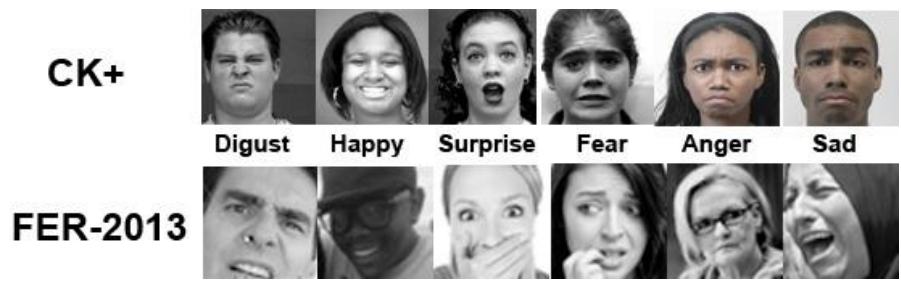

**Fig. 4.** Examples of CK+ and FER-2013 datasets

### 6.1 Experimental results for FER 2013

All the models are trained on 28709 examples in the FER 2013 dataset. The public set is used as validation set to tune the hyperparameters while the private set is used as test set. We initialized the weights as described in [6]. Each network is trained for 300 epochs with a batch size of 128.

Fig. 5a. shows the accuracy of the models on the test data. CNN with Dense SIFT outperforms the CNN only and the CNN with SIFT. Surprisingly the SIFT features did not improve the accuracy over the CNN only model. One possible reason is the SIFT Key-points locator is not suitable for 48x48 small images. In contrast, the Dense SIFT model increased the accuracy of the CNN by about 1%. Dense SIFT has another advantage in which it runs faster than regular SIFT as the key-points locations are already fixed and Bag of Key Points is not needed. The aggregator of the three models improves the result remarkably which surpasses state-of-the-art methods as shown in Table 2. Although the average sum technique is trivial and simple compared to [9], high variations in the model's designs increase the impact of the model's decisions.

### 6.2 Experimental Results for CK+ Dataset

OpenCV Cascade Classifier has been chosen to detect faces landmarks in the images in CK+ and these landmarks are used to crop the faces. The model is pre-trained on FER-2013 training set first and parameter tuning is performed on the CK+ dataset. The advantages of using the pre-trained model are to speed-up the training phase and to increase the overall performance of the model. We use all the 309 images in the dataset for training and testing is performed using 10-fold cross-validation. All the networks are trained only for 20 epochs to prevent overfitting due to the limited size of training set.

The addition of Dense SIFT to CNN has again shown better performance against CNN only and CNN with SIFT models as depicted in Fig. 5b. Both CNN with Dense SIFT model and the aggregator exhibit significant results as shown in Table 3. Moreover, all the models including the CNN only model outperforms state-of-the-art techniques. Several factors contribute to this good performance. For example, the eight layers in the deep neural network, extensive pre-processing, D-SIFT fusion with the CNN, and the aggregator of different models increase the discriminative power of the proposed method.

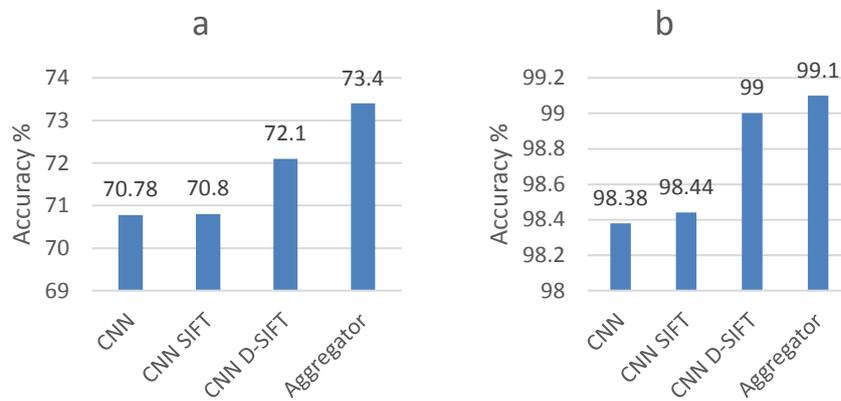

**Fig. 5.** Performance of the models on FER-2013 and CK

**Table 2.** Classification accuracies of different models on FER

| Method | Accuracy % |
|---|---|
| CNN SIFT | 70.8 |
| CNN D-SIFT | 72.1 |
| Proposed aggregator | **73.4** |
| (Kim et al., 2015) [9] | 72.72 |
| (Tang, 2013) [22] | 71.2 |
| (Mollahosseini et al., 2016) [15] | 66.4 |

**Table 3.** Classification accuracies of different models on CK+

| Method | Accuracy % |
|---|---|
| CNN only | 98.38 ± 0.03 |
| CNN SIFT | 98.44 ± 0.08 |
| CNN D-SIFT | 99 ± 0.07 |
| Proposed aggregator | **99.1 ± 0.07** |
| (Khorrami et al., 2015) [13] | 98.3 |

| (Lopes et al. 2017) [18] | 96.76 |
| (M. Liu, 2015) [12] | 93.70 |

## 7 Conclusion

In this paper, a hybrid Convolutional Neural Network with Dense Scale Invariant Feature Transform aggregator is proposed for facial expression recognition. We have shown how the Dense SIFT features and convolution neural network could complement each other in improving the accuracy result. The proposed method combines the strengths of hand-craft and deep learning approaches. The Dense SIFT technique is studied and compared with regular SIFT feature extractor in which it shows an advantage over regular SIFT. The performance gain is noticeable when all the models are combined with the aggregator in which it outperforms state-of-the-art methods. Our experiments demonstrate a clear advantage of aggregating Dense SIFT and CNN models by achieving outstanding results on both FER-2013 and CK+ datasets.